\documentclass[manuscript, screen]{acmart}
\AtBeginDocument{%
  }

\usepackage[hindi,english]{babel}
\usepackage{float}

\acmJournal{TALLIP}

\begin{document}

\title{Mitigating Structural Noise in Low-Resource S2TT: An Optimized
Cascaded Nepali-English Pipeline with Punctuation Restoration}

\author{Tangsang Chongbang}
\email{077bei047.tangsang@pcampus.edu.np}
\orcid{0009-0004-4221-5111}
\author{Pranesh Pyara Shrestha}
\email{077bei030.pranesh@pcampus.edu.np}

\author{Amrit Sarki}
\email{077bei049.amrit@pcampus.edu.np}

\author{Anku Jaiswal}
\email{anku.jaiswal@pcampus.edu.np}

\affiliation{%
  \institution{Department of Electronics and Computer Engineering, Institute of Engineering, Pulchowk Campus}
  \city{Lalitpur}
  \state{Bagmati}
  \country{Nepal}
}









\begin{abstract}
Cascaded speech-to-text translation (S2TT) systems for low-resource languages can suffer from structural noise, particularly the loss of punctuation during the Automatic Speech Recognition (ASR) phase. This research investigates the impact of such noise on Nepali-to-English translation and proposes an optimized pipeline to mitigate quality degradation. We first establish highly proficient ASR and NMT components: a Wav2Vec2-XLS-R-300m model achieved a state-of-the-art 2.72\% CER on OpenSLR-54, and a multi-stage fine-tuned MarianMT model reached a 28.32 BLEU score on the FLORES-200 benchmark. We empirically investigate the influence of punctuation loss, demonstrating that unpunctuated ASR output significantly degrades translation quality, causing a massive 20.7\% relative BLEU drop on the FLORES benchmark. To overcome this, we propose and evaluate an intermediate Punctuation Restoration Module (PRM). The final S2TT pipeline was tested across three configurations on a custom dataset. The optimal configuration, which applied the PRM directly to ASR output, achieved a 4.90 BLEU point gain over the direct ASR-to-NMT baseline (BLEU 36.38 vs. 31.48). This improvement was validated by human assessment, which confirmed the optimized pipeline's superior Adequacy (3.673) and Fluency (3.804) with inter-rater reliability (Krippendorff's $\alpha \geq 0.723$). This work validates that targeted punctuation restoration is the most effective intervention for mitigating structural noise in the Nepali S2TT pipeline. It establishes an optimized baseline and demonstrates a critical architectural insight for developing cascaded speech translation systems for similar low-resource languages. 

\end{abstract}

\ccsdesc[500]{Computing methodologies~Speech recognition}
\ccsdesc[500]{Computing methodologies~Machine translation}
\ccsdesc[300]{Computing methodologies~Language resources}
\ccsdesc[300]{Computing methodologies~Natural language generation}

\keywords{Speech-to-Text Translation, Low-Resource Languages, Nepali, ASR, NMT, Nepali-to-English, Punctuation Restoration, Wav2Vec2, MarianMT, Cascaded Pipeline.}


\maketitle

\section{Introduction}



Speech-to-text translation (S2TT) converts spoken speech from a source into written text in a target language. It is crucial for bridging communication barriers, enhancing digital accessibility and enabling applications like voice-command systems, automatic subtitling,  virtual assistants, and cross-lingual communication.
While Automatic Speech Recognition (ASR) and Machine Translation (MT) have seen significant advancements in high-resource languages, integrated S2TT systems remain underdeveloped and unexplored for low-resource languages (LRLs) like Nepali. 

Nepali, spoken by approximately 19 million native speakers, and an additional 14 million second-language speakers \cite{ethnologue2024}, presents a range of challenges for speech translation. These include limited annotated data, frequent code-switching with English \cite{gurung2019nepali}, regional dialect variation, and sociolinguistic features such as complex honorifics without direct English equivalents (e.g., {\dn t} vs. {\dn tpA\312w} both translate as ``you'' but differ in levels of formality). These characteristics complicate both ASR and MT for Nepali. 

End-to-end direct Speech-to-Speech translation (S2ST) models are gaining attention in recent years, for their advantages of architectural simplicity, and potential to reduce information loss, and error propagation \cite{bentivogli2021cascade}. However, such approaches demand large-scale, parallel speech and text data \cite{sarim2025direct}, which makes it impractical for Nepali and most LRLs. In contrast, cascaded pipelines, which first transcribe speech with ASR and then translate it with MT, offer a more practical alternative in low-resource settings. Their modularity allows optimization of each component, resulting in more powerful performance \cite{sarim2025direct}. Modern self-supervised ASR models like Wav2Vec2 have shown strong performance even with limited labeled data \cite{yi2020applying}, and multilingual NMT models like MarianMT and NLLB can be adapted even with modest amounts of LRL data to achieve good performance \cite{liu2020multilingual, verma2022strategies}. Consequently, for low-resource scenarios, cascaded pipelines offer greater data efficiency. 

However, cascaded systems introduce their own unique challenges. Error propagation is the most prominent example. A particularly underexplored issue in $\text{Nepali} \leftrightarrow \text{English}$ literature is the lack of punctuation in ASR outputs. While ASR models primarily focus on word accuracy, NMT models are trained on punctuated text, relying on punctuation marks for cues such as sentence boundaries, clause separation, emphasis, disambiguation. Their presence or absence can have significant impact on translation quality \cite{jwalapuram2023pulling}.

In this work, we develop and analyze cascaded $\text{Nepali} \rightarrow \text{English}$ S2TT system combining three main components: (1) a Wav2Vec2.0 XLS-R 300m ASR model (2) a Punctuation Restoration Module (PRM) using multilingual T5 (mT5) model for basic punctuation restoration, and (3) a multi-stage fine-tuned MarianMT NMT model. We confirm the component quality, with the ASR achieving a state-of-the-art 2.72\% CER on OpenSLR-54 test set and the NMT achieving 28.32 BLEU on the FLORES-200. Critically, we demonstrate that the absence of punctuation causes 20.7\% relative BLEU drop in translation quality, empirically validating the need for the PRM. The end-to-end evaluation shows that the PRM-optimized pipeline yields a 4.90 BLEU point gain over the direct ASR-to-NMT baseline. The key contributions of this work are as follows:
\begin{itemize}
    \item Establishing a new, optimized S2TT baseline for Nepali $\rightarrow$ English. We develop the first publicly documented cascaded Nepali $\rightarrow$ English S2TT system, demonstrating highly competitive component performance with a 2.72\% CER for ASR and a 28.32 BLEU score for NMT on standard benchmarks. 
    \item  Validating Punctuation Restoration as the Optimal Intervention. We quantitatively demonstrate that the loss of punctuation leads to a 20.7\% relative BLEU drop, and show that incorporating an intermediate PRM yields robust 4.90 BLEU point gain in the end-to-end pipeline. 
\end{itemize}

The rest of the paper is structured as follows: Section 2 reviews related work.
Section 3 outlines the methodology, including data preparation and experimental
setups. Section 4 presents results and findings from the component-level
evaluations, the end-to-end S2TT scenarios and the human evaluation results.
Section 5 discusses the findings, and Section 6 finally concludes the paper.

\section{Related Work}

This section reviews prior research in ASR, NMT, and their integration into S2TT pipelines. We specifically highlight the unique challenges and existing gaps in the literature concerning Nepali, for which a unified S2TT system has yet to be publicly documented. 

\subsection{Automatic Speech Recognition (ASR) for Low-Resource Languages}

ASR in low-resource settings suffers from labeled data scarcity, diverse dialects and frequent code-switching. 

Traditional ASR approaches for Nepali relied on Hidden Markov Models (HMMs), as in the early work by \cite{ssarma2017hmm}, which achieved 74.49\% accuracy for single-word inputs and 55.55\% for three-word phrase inputs Nepali recognition using Voice Activity Detection (VAD), primarily limited by their ability to model complex acoustic variations effectively.
The advent of deep learning introduced more sophisticated models like RNN-CTC and CNN-RNN hybrids. \cite{regmi2019nepali} implemented an RNN-CTC model trained on a 1,320-word custom dataset, that achieved a 34\% CER, but struggled with speaker variability and generalization. \cite{bhatta2020nepali} reported 1.83\% CER and 11\% WER using a CNN-GRU architecture on the OpenSLR43 text-to-speech (TTS) dataset. However, the studio-quality nature of OpenSLR43 limits real-world applicability. Other efforts, such as \cite{dhakal2022automatic}, utilized bidirectional LSTM paired with  ResNet and one-dimensional CNN to achieve a 17.06\% CER on OpenSLR dataset. While these studies laid foundational groundwork for Nepali ASR, the models were less capable of generalizing to natural speech variations. 

Recent self-supervised models like Wav2Vec2 and Whisper have reshaped ASR for LRLs. By leveraging large-scale unlabeled data for pre-training, they enable better generalization with limited examples \cite{zhu2021wav2vec, hsu2024meta, fatehiself}. Our work builds upon this paradigm by utilizing the Wav2Vec2-XLS-R-300m model on OpenSLR54 and Common Voice v17. This approach enables us to establish a competitive performance benchmark for Nepali ASR.

\subsection{Machine Translation (MT) for Low-Resource Languages}
Similar to ASR, NMT for LRLs is constrained by lack of high-quality parallel corpora and linguistic complexities. Traditional Statistical Machine Translation (SMT) and rule-based methods dominated early efforts, but transformer-based NMT has since become the standard. However, NMT models require large datasets and are sensitive to rare words, long sentences, domain mismatch and word-alignment issues \cite{koehn2017six}. 

To address these limitations, researchers have explored transfer learning, backtranslation, data augmentation, and pivot languages \cite{haddow2022survey, talwar2025pivot, zoph2016transfer}. For instance, MarianMT and NLLB are pretrained multilingual models that support rapid adaptation to LRLs, even with modest amounts of data \cite{liu2020multilingual, verma2022strategies}. Recent work by \cite{verma2022strategies} demonstrated the effectiveness of multi-stage fine-tuning strategy involving multilingual pre-training followed by language-pair specific fine-tuning, followed by domain fine-tuning. 

Inspired by these successes, our approach applies a similar multi-stage fine-tuning strategy to a MarianMT mul-en model using filtered NLLB data, synthetic parallel corpora, and a high-quality dataset to achieve substantial improvements in translation quality for $\text{Nepali} \rightarrow \text{English}$.

\subsection{Challenges in ASR-MT Integration and Punctuation Restoration}

Cascaded ASR-MT pipelines, while practical for LRLs, are inherently susceptible to error propagation, where transcription errors negatively affect downstream translation. Since ASR models like Wav2Vec2 primarily focus on word accuracy, and are not inherently trained to predict punctuation, the resultant degradation from directly feeding the unpunctuated ASR output to NMT models is an under-explored issue, especially for LRLs. Punctuation conveys structural and semantic cues like sentence boundaries, emphasis, disambiguation, which NMT models heavily rely on for accurate translation. The work by \cite{jwalapuram2023pulling} quantified the effects of punctuation on MT specifically for German-English, Japanese-English and Ukrainian-English language pairs and concluded that models are heavily sensitive to punctuation. 

While prior works in high-resource languages have addressed punctuation
restoration as a standalone task or via joint training, its role in ASR-MT
pipelines for Nepali remains unexplored. Our work explicitly quantifies the
degradation in translation performance due to the absence of punctuation,
thereby underscoring the necessity for an intermediate structural noise
mitigation component. Based on this gap, Section 3 details the comprehensive methodology developed to create and evaluate an optimized, punctuation-aware cascaded Nepali $\rightarrow$ English S2TT system.

\section{Methodology}
This section details the design, implementation, and training procedures of our cascaded Nepali S2TT system. We first outline the overall system architecture. This is followed by a comprehensive description of the datasets used and the fine-tuning strategies applied to the ASR and NMT components. We then describe the implementation of the preliminary punctuation restoration module and define the specific evaluation scenarios employed.

\subsection{Overall System Architecture}

\begin{figure}[H]
   \includegraphics[width=\textwidth]{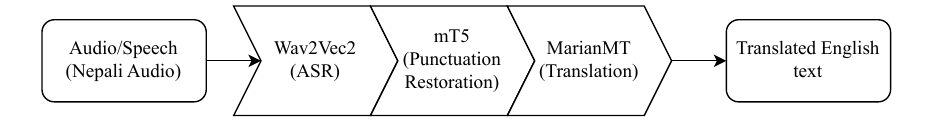}
   \caption{Cascaded Nepali-to-English Speech-To-Text-Translation System Architecture}
   \label{fig:system_architecture}
 \end{figure}

The proposed system adopts a cascaded architecture (Figure \ref{fig:system_architecture}) consisting of three sequential stages.
\begin{itemize}
    \item \textbf{Stage 1 - ASR Transcription:} Nepali speech is processed by the fine-tuned Wav2Vec2-XLS-R-300m model to generate an unpunctuated Nepali text transcription.
    \item \textbf{Stage 2 - Punctuation Restoration (Conditional):} The unpunctuated ASR output is passed through a preliminary mT5 model for the restoration of basic punctuation, primarily sentence-ending full stops. This stage is used only in evaluation scenarios to quantify the impact of punctuation recovery.

    \item \textbf{Stage 3 - NMT Translation:} The transcribed text is fed into the fine-tuned MarianMT mul-en model, which produces the final English text translation. 
\end{itemize}

 \subsection{Datasets}
This section details the construction and preprocessing of the main corpora utilized in this study: two for ASR fine-tuning, three for NMT multi-stage training, one for punctuation restoration, and one for the specialized end-to-end evaluation.

\subsubsection{ASR Datasets}
The ASR model was fine-tuned on a combination of two publicly available Nepali speech corpora to increase speaker diversity and utterance coverage (see Table \ref{tab:asr-speech-datasets}).

\begin{table}[ht]
  \centering
  \small
  \setlength{\tabcolsep}{3pt} 
  \renewcommand{\arraystretch}{1.2}
  \caption{Nepali ASR datasets used for fine-tuning.}
  \label{tab:asr-speech-datasets}
  
  \begin{tabular}{p{3cm} p{6cm} c c}
    \toprule
    \textbf{Dataset} & \textbf{Total Utterances/Clips} & \textbf{Hours (Validated)} & \textbf{Used Utterances} \\
    \midrule
    Common Voice v17\textsuperscript{a} & 742 (validated) + 595 (internally reviewed `other' split) = 1,337 &
    1.0 & 1,337 \\
    
    OpenSLR54\textsuperscript{b} & $\sim$157{,}000 (initial) $\rightarrow$ 148{,}188 (post-numerals-filtering) &
    $\sim$143 (pre-5s filter) & 136,095 (post-5s filter) \\
    \bottomrule
  \end{tabular}
  
  \vspace{1ex}
  \raggedright 
  \footnotesize
  \textsuperscript{a} \url{https://huggingface.co/datasets/mozilla-foundation/common_voice_17_0} \\
  \textsuperscript{b} \url{https://huggingface.co/datasets/iamTangsang/OpenSLR54-Nepali-ASR}
\end{table}

From the Common Voice v17 (ne-NP) dataset, we used 1,337 utterances (approximately 2 hours) from 32 speakers. We combined the 'validated' split with internally reviewed and validated samples from the 'other' split for initial fine-tuning. All audio files were uniformly resampled to 16 kHz.

For OpenSLR54, utterances containing numerals were removed to ensure the model outputs numbers in words rather than digits, since inconsistent transcriptions were common. For example, the numeral {\dn 007} appeared as "{\dn Ejro Ejro sAt}" in one recording, "{\dn  \8{s}\306wy \8{s}\306wy sAt}" in another, and simply "{\dn sAt}" elsewhere. To avoid confusing the model, all such utterances were excluded. 

In addition, since sequences longer than 5 seconds caused excessive processing time and frequent out-of-memory errors on the Colab environment, these were also removed. After filtering, 136,095 utterances remained for final fine-tuning.

\subsubsection{NMT Datasets and Multi-Stage Strategy}

The Nepali-to-English NMT model was fine-tuned using a three-stage approach. This strategy leverages broad language knowledge from noisy sources first, expands coverage with high-volume synthetic data, and concludes with a refinement stage using high-fidelity data (see Table \ref{tab:training-stages}).
\begin{table}[ht]
  \small
  \centering
  \caption{Corpora used at each stage of NMT training and their filtering/rationale.}
  \label{tab:training-stages}
  
  \setlength{\tabcolsep}{5pt}
  \begin{tabular}{p{1.4cm} p{2cm} r p{4.3cm} p{4.3cm}}
    \toprule
    \textbf{Stage} & \textbf{Corpus Type} & \textbf{Size (Pairs)} & 
    \textbf{Primary Source / Filtering} & \textbf{Rationale} \\
    \midrule
    Stage 1: Foundational & Filtered Web crawled NLLB\textsuperscript{a} & 708{,}000 & Initial 19.6M web-crawled NLLB corpus pairs filtered by LabSE cosine similarity ($>0.80$), and removal of Nepali numerals (to maintain consistency). &
    Knowledge Transfer: To learn fundamental translation nuances from filtered, noisy web data. \\
    \midrule
    
    Stage 2: Expansion & Synthetic Corpora\textsuperscript{b,c} & 5.02M &
    Combination of 1.6M synthetic pairs \cite{duwal2019efforts} and 3.42M pre-training pairs \cite{duwal2024domain} filtered using a chrF++ cut-off of 50 (translated with 8-bit NLLB/IndicTrans2). &
    Coverage Expansion: Increase exposure to diverse structures and vocabularies using high-volume, quality-filtered synthetic data. \\

    \midrule
    
    Stage 3: Refinement & Manually translated high-Quality corpora\textsuperscript{d,e} & 210{,}875 &
    OPUS (GNOME/KDE/Ubuntu), Bible, Global Voices, Penn Treebank + \cite{acharya2018comparative}, Nepal Law Commission, Easy Bible, Nepal Budget Speech 2081/82, Shirish ko Fool Translation. &
    Quality Refinement: Final tuning for translation fluency and accuracy on high-fidelity human-translated pairs. \\

    \bottomrule
  \end{tabular}
  
  \vspace{1ex}
  \raggedright 
  \footnotesize
  \textsuperscript{a} \url{https://huggingface.co/datasets/iamTangsang/Nepali-to-English-Translation-Dataset} \\
  \textsuperscript{b} \url{https://huggingface.co/datasets/sharad461/ne-en-synthetic-1.6m} \\
  \textsuperscript{c} 
  Dataset described in \cite{duwal2024domain}; currently unavailable online. \\
  \textsuperscript{d} \url{https://huggingface.co/datasets/sharad461/ne-en-parallel-177k} \\
  \textsuperscript{e} \url{https://github.com/BISHALTWR/Nepali-English-Translation-Dataset}
\end{table}

\subsubsection{Punctuation and Segmentation Dataset}
\label{subsec: punctuation-dataset}
The punctuation Restoration module was trained to simultaneously address two critical issues observed in the ASR output: missing punctuation and fused words (lack of segmentation).

The training corpus was derived from the 210,875 high-quality sentence pairs from Stage 3 of the NMT training.
\begin{itemize}
  \item \textbf{Input Sequence:} The original Nepali sentences were systematically modified under two distinct conditions: (a) by removing both inter-word spaces and punctuation marks to produce completely fused character sequences  (e.g., {\dn dAUdl\?(yss\4EnklAIBn\?EtmFl\?ksrFjA\306wyOfAUlrjonATnmr\?ko\7{k}rA}, and (b) by removing only punctuation marks while preserving boundaries. These modifications were designed to emulate the types of errors commonly encountered in our ASR outputs. 
  \item \textbf{Target Sequence:} The corresponding reference Nepali text, fully segmented and punctuated, representing the desired restoration output (e.g., {\dn dAUdl\? (ys s\4EnklAI Bn\?{\rs ,"\re} EtmFl\? ksrF jA\306wyO fAUl r jonATn mr\?ko \7{k}rA{\rs ?"\re}}) 
\end{itemize}
The final dataset comprised a combination of both modification types, enabling the model to learn from varying degrees of degradation. By framing this as a unified text-to-text generation task, the mT5 model was trained to jointly recover accurate word boundaries and restore essential punctuation marks (commas, and full stops) in a single step.

\subsubsection{Final End-to-End Evaluation Dataset}
The final performance of the full cascaded system was assessed on a newly created, representative test set\footnote{The final evaluation dataset can be accessed through Hugging Face at \url{https://huggingface.co/datasets/iamTangsang/nepali_to_english_pipeline_evaluation}} to simulate real-world usage.
The final performance of the full cascaded system was assessed on a newly created, representative test set to simulate real-world usage.

This evaluation dataset consists of 900 total audio clips, comprising 300 unique Nepali sentences recorded by 3 different speakers.

  \paragraph{Sentence Selection:} The 300 sentences were manually crafted to ensure linguistic diversity, including: \begin{itemize}
    \item Statements (150): Simple declarative sentences.
    \item Questions (60): To test the model's ability to recognize interrogative punctuation.
    \item Commands/Imperatives (30): To test varied sentence structures.
    \item Complex/Compound Sentences (30): To test handling of multiple clauses.
    \item Named Enities (30): To test proper noun transcription and translation.
  \end{itemize}
  \paragraph{Annotation:} All audio was manually transcribed, punctuated, and translated to establish the Gold-Standard reference used for final S2TT evaluation. 

\subsection{Model Training and Fine-tuning}
All models were trained and fine-tuned using the Google Colab environment on a Tesla T4 GPU. The training strategies were modularly defined for the ASR, NMT, and Punctuation Restoration components to ensure optimal performance in the low-resource setting.

\subsubsection{ASR Model Fine-tuning}
The Wav2Vec2.0 XLS-R model (300 million parameters variant) was selected as the foundational architecture. The fine-tuning process was executed in a three-stage sequential manner, incorporating iterative vocabulary and learning rate optimization.
\paragraph{Initial Tuning on Common Voice v17:}
The model was initially fine-tuned on a combined set of 1,337 items from the 'validated' and internally reviewed 'other' splits of the Common Voice v17 ne-NP corpus.
\begin{itemize}
  \item Training was performed for 30 epochs with a learning rate of 3e-4.
  \item Vocabulary: The initial vocabulary size was of 64 tokens.
\end{itemize}

\paragraph{Fine-tuning on OpenSLR 54 (Stage 1 - Plateau):}
The model obtained from CV-17 was further fine-tuned on the OpenSLR54 dataset.
\begin{itemize}
  \item \textbf{Initial Parameters:} The learning rate was set to 3e-4 with a linear
  decay for 16 epochs. A large, exploratory vocabulary of 71 tokens (including
  special tokens \texttt{<s>}, \texttt{</s>}, \texttt{\_\_UNK\_\_},
  \texttt{\_\_PAD\_\_}) was utilized.
  
  \item \textbf{Observation:} Training plateaued around the $16^{\text{th}}$ epoch (WER
  fluctuating 26\%-29\%).
\end{itemize}

\paragraph{Fine-tuning on the OpenSLR54 (Stage 2 - Refinement and Optimization):}
To address the plateauing, the model was retrained for an additional 3 epochs using key optimizations:
\begin{itemize}
  \item \textbf{Learning Rate:} Reduced significantly to 2e-5 with linear decay.
  \item \textbf{Vocabulary Optimization:} The final vocabulary size was reduced to 67 total tokens by removing characters not used in standard Nepali, which improved model stability and generalization.
\end{itemize}

\paragraph{Model Regularization:}
The Wav2Vec2 XLS-R utilized the following regularization parameters:
attention\_dropout\: 0.1, hidden\_dropout\: 0.1, layerdrop: 0.1. The CTC loss reduction was set to "mean". The key hyperparameters used were as shown in Table \ref{tab:hyperparameters}.
\begin{table}[ht]
  \centering
  \caption{Key Hyperparameters for Wav2Vec2.0 Fine-Tuning}
  \label{tab:hyperparameters}
  
  \setlength{\tabcolsep}{10pt} 
  \begin{tabular}{l l}
    \toprule
    \textbf{Hyperparameter} & \textbf{Value} \\
    \midrule
    Learning Rate (Final) & $2 \times 10^{-5}$ \\
    Batch Size (Per Device) & 16 \\
    Gradient Accumulation & 2 \\
    Warmup Steps & 500 \\
    \bottomrule
  \end{tabular}
\end{table}

\subsubsection{Punctuation Restoration Module Fine-tuning}
The preliminary Punctuation Restoration module employed the Google mT5 model fine-tuned on the dataset defined in Section~\ref{subsec: punctuation-dataset}. The training was framed as a single text-to-text task to simultaneously restore segmentation and punctuation. The key hyperparameters used were as shown in Table \ref{tab:mt5_hyperparameters}.
\begin{table}[ht]
  \centering
  \caption{Key Hyperparameters for mT5 Fine-Tuning}
  \label{tab:mt5_hyperparameters}
  
  \setlength{\tabcolsep}{10pt} 
  \begin{tabular}{l l}
    \toprule
    \textbf{Hyperparameter} & \textbf{Value} \\
    \midrule
    Learning Rate & $2 \times 10^{-5}$ \\
    Batch Size (Train) & 8 \\
    Weight Decay & 0.1 \\
    \bottomrule
  \end{tabular}
\end{table}

\subsubsection{NMT Model Multi-Stage Fine-tuning}
The MarianMT mul-en model underwent a sequential, multi-stage fine-tuning process to effectively utilize the increasingly high-quality data corpora.
\paragraph{Stage 1 \& 2: Pre-training (NLLB Filtered \& Synthetic Datasets):}
The model was initially trained on the 708,000 filtered NLLB corpus (Stage 1) for a single epoch to establish foundational transfer knowledge. Training then continued on the combined 5.02 million synthetic corpora (Stage 2) to expand linguistic coverage.
\begin{itemize}
  \item Learning Rate: 5e-5 was used across the pre-training stages for a balance between speed and stability.
  \item Token Limits: The input and output were capped at 256 tokens.
\end{itemize}
\paragraph{Stage 3: Fine-tuning (High-Quality Dataset):}
The model was finally refined on the 210,875 high-quality parallel pairs for 12 epochs. Table \ref{tab:marianmt_hyperparams} summarizes the hyperparameters used.
\begin{table}[ht]
  \centering
  \caption{Hyperparameters for MarianMT Pre-training and Fine-tuning}
  \label{tab:marianmt_hyperparams}
  
  \setlength{\tabcolsep}{6pt} 
  \begin{tabular}{l l l}
    \toprule
    \textbf{Hyperparameter} & \textbf{Pre-training} & \textbf{Fine-tuning} \\
    \midrule
    Learning Rate & $5 \times 10^{-5}$ & $2 \times 10^{-5}$ \\
    Batch Size (Train/Eval) & 16 & 16 \\
    Weight Decay & 0.01 & 0.01 \\
    Warmup Steps & 5\% of total steps & 5\% of total steps \\
    \bottomrule
  \end{tabular}
\end{table}

\subsection{Evaluation Metrics}
Performance was measured using task-specific metrics as in Table \ref{tab:evaluation_types}.
\begin{table}[ht]
  \centering
  \caption{Evaluation Types and Corresponding Metrics}
  \label{tab:evaluation_types}
  
  \setlength{\tabcolsep}{10pt} 
  \begin{tabular}{p{4cm} l}
    \toprule
    \textbf{Evaluation Type} & \textbf{Metrics} \\
    \midrule
    ASR & WER, CER \\
    NMT Benchmarking & BLEU, chrF++, METEOR \\
    Punctuation Impact \& S2TT & $\Delta$BLEU, $\Delta$chrF++ \\
    Human Evaluation & 1--5 Likert Scale \\
    \bottomrule
  \end{tabular}
\end{table}

\subsection{Experimental Setup and Evaluation Scenarios}
The system performance was rigorously evaluated across three core experimental setups:
\begin{itemize}
  \item[1.] \textbf{Component Performance:} Individual evaluation of ASR (on OpenSLR-54 test set split) and NMT (on FLORES-200/Tatoeba and the high-quality test split) to establish component-level baselines.
  \item[2.] \textbf{Punctuation Impact Quantification:} \begin{itemize}
    \item[•] \textbf{Procedure:} The NMT model was evaluated on the FLORES-200 dev, devtest, and Tatoeba datasets. This was done by comparing the performance on (1) the original, fully punctuated source sentences against (2) a modified version of the same sentences with all punctuation removed (Simulated ASR output).
    \item[•] \textbf{Goal:} The resulting perforamnce delta ($\Delta$ BLEU and $\Delta$ chrF++) quantifies the raw performance degradation directly attributable to the loss of source-side punctuation.  
  \end{itemize} 
  \item[3.] \textbf{End-to-End (S2TT) Evaluation:} The full cascaded pipeline was tested on the 900-clip custom dataset across three scenarios: \begin{itemize}
    \item[•] \textbf{Scenario A (Direct Baseline):} ASR(unpunctuated output) $\rightarrow$ NMT.
    \item[•] \textbf{Scenario B:} ASR $\rightarrow$ Punctuation Restoration Module (with space removal from ASR's output) $\rightarrow$ NMT.
    \item[•] \textbf{Scenario C:} ASR $\rightarrow$ Punctuation Restoration Module (without space removal from ASR's output) $\rightarrow$ NMT. 
  \end{itemize}
\end{itemize}
\paragraph{Human Evaluation (S2TT):} A dedicated human evaluation was conducted across all three S2TT scenarios using the 900-clip custom test set. Three of the authors conducted a blind human evaluation to assess translation fluency and adequacy on a 1 -- 5 Likert scale. The outputs from all system configurations were randomly mixed and anonymized prior to scoring, ensuring that evaluators were unaware of which system configuration produced each translation. After all the ratings were completed, the configuration identities were revealed for analysis. Scores were averaged across raters for each system. To validate the consistency of the human assessments, inter-rater reliability was calculated using Krippendorff's Alpha($\alpha$) for both metric, treating the Likert scores as ordinal data.

\section{Results and Findings}
This section presents the empirical results of the cascaded S2TT system, structured arond the performance of its components, the quantified impact of noise, and the final end-to-end efficacy of the proposed pipeline.
\subsection{Component-Level Performance}
\subsubsection{ASR Performance}
The fine-tuned Wav2Vec2 XLS-R model was evaluated on the OpenSLR-54 test set (see Table \ref{tab:asr_performance}). The resulting WER of 16.82\% and a particularly low CER of 2.72\% confirm the viability of the self-supervised transformer approach for low-resource Nepali ASR. This CER represents a substantial improvement over previous established supervised models on the same dataset (e.g., \cite{dhakal2022automatic} at 17.07\% CER and \cite{banjara2020nepali} at 27.72\% CER).
\begin{table}[ht]
  \centering
  \caption{ASR Performance on OpenSLR-54 Test Set}
  \label{tab:asr_performance}
  
  \setlength{\tabcolsep}{10pt} 
  \begin{tabular}{p{5cm} l}
    \toprule
    \textbf{Metric} & \textbf{Result} \\
    \midrule
    Word Error Rate (WER) & 16.82\% \\ 
    Character Error Rate (CER) & 2.72\% \\
    \bottomrule
  \end{tabular}
\end{table}
\subsubsection{NMT Benchmarking}
The final NMT model was benchmarked against widely used standard sets and the base model. The results are reported in Table \ref{tab:marianmt_benchmark}.
\begin{table}[ht]
  \centering
  \caption{Benchmarking of the Final MarianMT Model Against Standard Test Sets and Previous Best Results}
  \label{tab:marianmt_benchmark}
  
  \setlength{\tabcolsep}{4pt} 
  \begin{tabular}{l l r r}
    \toprule
    \textbf{Test Set} & \textbf{Metric} & \textbf{Our Score} & \textbf{Previous Best} \\
    \midrule
    FLORES-200 DevTest & BLEU   & \textbf{29.04} & 20.76 \\
                       & chrF++ & \textbf{58.14} & N/A \\
                       & METEOR & \textbf{0.6314} & N/A \\
    \midrule
    FLORES-200 Dev     & BLEU   & \textbf{28.48} & 19.37 \\
                       & chrF++ & \textbf{58.07} & N/A \\
                       & METEOR & \textbf{0.5676} & N/A \\ 
    \midrule
    Tatoeba            & BLEU   & \textbf{39.66} & 3.5\textsuperscript{a} \\
                       & chrF++ & \textbf{55.73} & 0.168\textsuperscript{a} \\
                       & METEOR & \textbf{0.5532} & N/A \\
    \bottomrule
  \end{tabular}
  
  \vspace{1ex}
  \raggedright 
  \footnotesize
  \textsuperscript{a} Base MarianMT mul-en: \href{https://huggingface.co/Helsinki-NLP/opus-mt-mul-en}{https://huggingface.co/Helsinki-NLP/opus-mt-mul-en}.
\end{table}
The NMT model achieved significant gains across all benchmarks, surpassing the previously reported state-of-the-art results \cite{duwal2019efforts} on FLORES-200 by $\sim 8$--$9$ BLEU points and drastically outperforming the base multilingual MarianMT model. This performance ensures that the NMT component is highly proficient when provided with high-quality, fully punctuated source text.

\subsection{Quantifying Punctuation Impact on NMT}
\begin{figure}[ht]
   \includegraphics[width=0.9\textwidth]{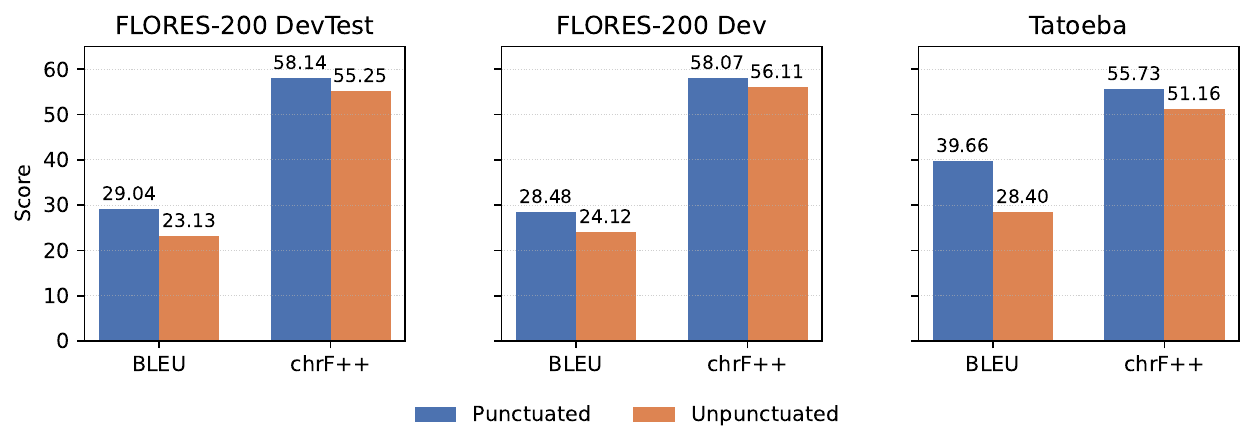}
   \caption{Comparative performance of NMT on punctuated and unpunctuated (Simulated ASR) inputs across three test sets. A consistent degradation is observed, confirming that punctuation loss substantially impacts translation quality.}
   \label{fig:punct-impact}
 \end{figure}

To isolate the impact of noise common to ASR output, the NMT model was tested in two modes: with and without Nepali punctuation. Quantitative results are reported in Table \ref{tab:punctuation_impact} and visualized in Figure \ref{fig:punct-impact}. The results confirm punctuation removal leads to notable degradation in translation performance across all datasets. Notably, FLORES-200 DevTest exhibits a \textbf{5.91} BLEU drop (20.3\% relative), while Tatoeba suffers the most extreme degradation, exhibiting \textbf{11.26} BLEU drop (28.39\% relative). This empirical evidence provides the primary justification for integrating a Punctuation Restoration module to mitigate error propagation.
\begin{table}[ht]
  \centering
  \caption{Impact of Punctuation on Translation Quality Across Test Sets (MarianMT Model Evaluation)}
  \label{tab:punctuation_impact}
  
  \setlength{\tabcolsep}{4pt} 
  \begin{tabular}{l l r r r r}
    \toprule
    \textbf{Test Set} & \textbf{Condition} & \textbf{BLEU} & \textbf{chrF++} & \textbf{$\Delta$BLEU} & \textbf{$\Delta$chrF++} \\
    \midrule
    FLORES-200 DevTest & Punctuated & 29.04 & 58.14 & --- & --- \\
                       & Unpunctuated (ASR) & 23.13 & 55.25 & \textbf{-5.91} & \textbf{-2.89} \\[3pt]
    \midrule
    FLORES-200 Dev     & Punctuated & 28.48 & 58.07 & --- & --- \\
                       & Unpunctuated (ASR) & 24.12 & 56.11 & \textbf{-4.36} & \textbf{-1.97} \\[3pt]
    \midrule
    Tatoeba            & Punctuated & 39.66 & 55.73 & --- & --- \\
                       & Unpunctuated (ASR) & 28.40 & 51.16 & \textbf{-11.26} & \textbf{-4.57} \\
    \bottomrule
  \end{tabular}
\end{table}
 
 \subsection{End-to-End(S2TT) Evaluation}
The full cascaded system was evaluated on the 900-clip custom test set.
\subsubsection{Automatic Metrics (BLEU and chrF++)}
\begin{figure}[ht] 
\centering
\includegraphics[width=0.9\textwidth]{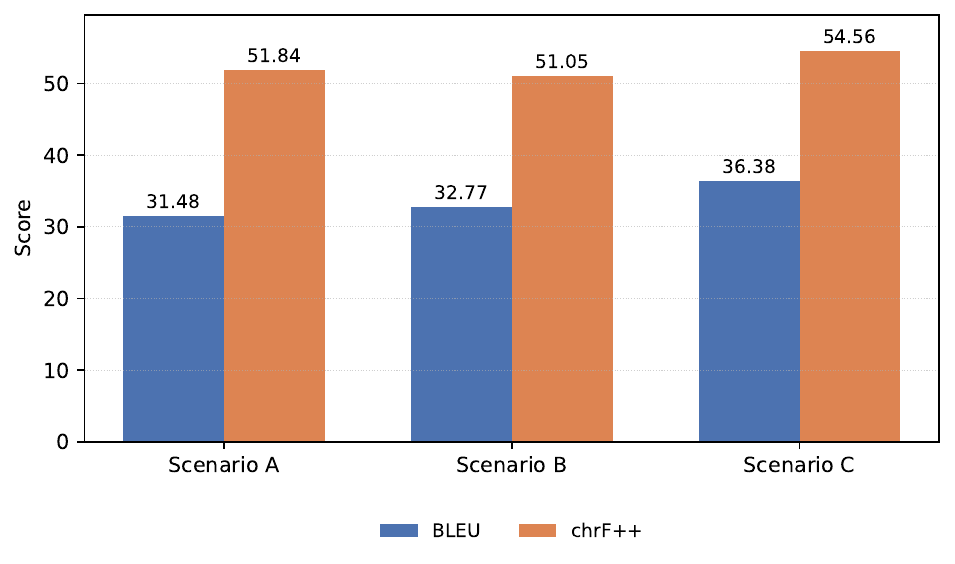}
\caption{End-to-End S2TT Performance Comparison (BLEU \& chrF++)}
\label{fig:scenario-barchart}
\end{figure}

The raw ASR output on this custom set had an average WER of \textbf{37.36\%} and CER of \textbf{13.76\%}, reflecting the difficulty in accuracy of the ASR model in real-world usage despite good performance in the OpenSLR54 test set. The three scenarios tested the efficacy of the Punctuation Restoration module. Table \ref{tab:system_config_impact} and Figure \ref{fig:scenario-barchart} summarizes the result.
\begin{table}[ht]
  \centering
  \caption{Impact of System Configuration on Translation Quality}
  \label{tab:system_config_impact}
  
  \setlength{\tabcolsep}{8pt} 
  \begin{tabular}{l c c r}
    \toprule
    \textbf{Scenario} & \textbf{BLEU} & \textbf{chrF++} & \textbf{$\Delta$BLEU (vs. Baseline A)} \\
    \midrule
    A (Baseline) & 31.48 & 51.84 & --- \\
    B (Proposed - Full Pipeline) & 32.77 & 51.05 & \textbf{+1.29} \\
    C (Optimal Pipeline)  & 36.38 & 54.56 & \textbf{+4.90} \\
    \bottomrule
  \end{tabular}
  
  \vspace{1ex}
  \raggedright 
  \footnotesize
  Scenario A: ASR $\rightarrow$ NMT\\
  Scenario B: ASR $\rightarrow$ Punctuation Restoration with space removal $\rightarrow$ NMT \\
  Scenario C: ASR $\rightarrow$ Punctuation Restorationwithout space removal $\rightarrow$ NMT.
\end{table}

Scenario C (Punctuation Restoration on the raw ASR output (without removing spaces)) yielded the most significant performance gain, achieving a \textbf{4.90 BLEU} point increase and a \textbf{2.72 chrF++} point increase over the direct baseline (Scenario A). The minor drop in performance of performance B suggests that the initial space removal and re-tokenization introduced more detrimental segmentation errors than it resolved. Or, it may be because of lack of enough data as we used only $\sim 208k$ examples to train our mT5 model. \textbf{Scenario C}, which effectively only restored punctuation on raw ASR output, is the optimal system configuration.

\subsubsection{Human Evaluation (Adequacy and Fluency)}
\begin{figure}[H] 
\centering
\includegraphics[width=0.9\textwidth]{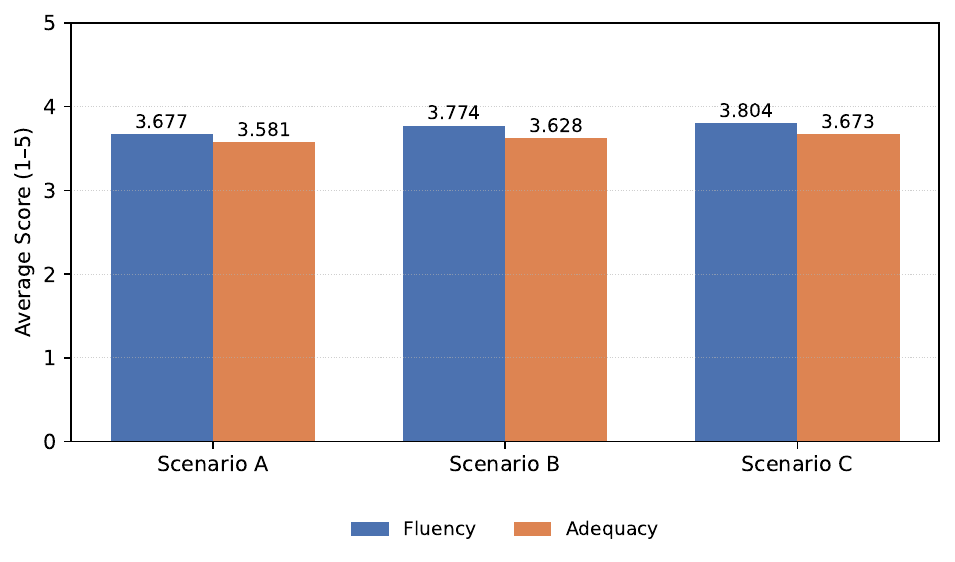}
\caption{Human evaluation of S2TT output. Average fluency and adequacy scores (1–5 scale) for Scenarios A, B, and C.}
\label{fig:human-eval-barchart}
\end{figure}

The human evaluation results, summarized in Table \ref{tab:system_config_fluency_adequacy} and illustrated in Figure \ref{fig:human-eval-barchart}, confirm the trends observed in the automatic evaluation metrics.
  
  

\begin{table}[H]
  \centering
  \caption{Impact of System Configuration on Translation Fluency and Adequacy with Inter-Rater Reliability ($\alpha$)}
  \label{tab:system_config_fluency_adequacy}
  
  \setlength{\tabcolsep}{6pt} 
  \begin{tabular}{l c c c c}
    \toprule
    \textbf{Scenario} & \textbf{Avg. Fluency} & \textbf{$\alpha$ (Fluency)} & \textbf{Avg. Adequacy} & \textbf{$\alpha$ (Adequacy)} \\
    \midrule
    A (Baseline) & 3.677 & 0.723 & 3.581 & 0.783 \\
    B (Proposed) & 3.774 & 0.746 & 3.628 & 0.780 \\
    C (Optimal)  & \textbf{3.804} & \textbf{0.748} & \textbf{3.673} & \textbf{0.785} \\
    \bottomrule
  \end{tabular}
  
  \vspace{1ex}
  \raggedright 
  \footnotesize
  Scenario A: ASR $\rightarrow$ NMT. \\
  Scenario B: ASR $\rightarrow$ Punctuation Restoration (with space removal) $\rightarrow$ NMT. \\
  Scenario C: ASR $\rightarrow$ Punctuation Restoration (without space removal) $\rightarrow$ NMT.
\end{table}

\textbf{Scenario C} consistently yields the highest-quality output, achieving the best scores for both \textbf{Fluency (3.804)} and \textbf{Adequacy (3.673)}. This result validates that the BLEU/chrF++ gains translate directly into a perceptibly higher-quality, more readable, and more meaning-preserving translations. Inter-rater reliability was measured using Krippendorff's Alpha($\alpha$) for both metrics, yielding high agreement ($\alpha \geq 0.723$) across all  systems, reinforcing the reliability of these findings.

\subsubsection{Performance Analysis by Sentence Type}
The breakdown by sentence category further highlights the impact of the punctuation restoration on specific linguistic structures. Table \ref{tab:sentence_type_eval} summarizes the result.

\begin{table}[ht]
  \centering
  \caption{Sentence-Type-wise Human Evaluation Results}
  \label{tab:sentence_type_eval}
  
  \setlength{\tabcolsep}{3pt} 
  \begin{tabular}{l l r r r}
    \toprule
    \textbf{Sentence Type} & \textbf{Metric} & \textbf{Scenario A (Baseline)} & \textbf{Scenario B} & \textbf{Scenario C (Optimal)} \\
    \midrule
    Statements      & Adequacy & 3.943 & 3.839 & \textbf{3.962} \\
    Statements      & Fluency  & 3.967 & 3.921 & \textbf{4.048} \\
    Commands        & Adequacy & 3.167 & \textbf{3.421} & 3.416 \\
    Commands        & Fluency  & 2.981 & 3.171 & \textbf{3.226} \\
    Questions       & Adequacy & 3.365 & 3.368 & \textbf{3.538} \\
    Questions       & Fluency  & 3.599 & 3.649 & \textbf{3.781} \\
    Named Entities  & Adequacy & 3.269 & \textbf{3.887} & 3.269 \\
    Named Entities  & Fluency  & 3.259 & \textbf{3.921} & 3.323 \\
    Complex         & Adequacy & 3.181 & 3.257 & \textbf{3.438} \\
    Complex         & Fluency  & 3.416 & 3.479 & \textbf{3.611} \\
    \bottomrule
  \end{tabular}
  
  \vspace{1ex}
  \raggedright 
  \footnotesize
  Evaluation by sentence category based on output type. Scores are averaged over three human evaluators using a 1--5 Likert scale. Higher scores indicate better perceived fluency or adequacy.
\end{table}

While Scenario C generally performs the best, \textbf{Scenario B} provided a superior outcome for the translation of Named Entities (Adequacy \textbf{3.887} vs. 3.269 and Fluency \textbf{3.921} vs. 3.323), suggesting that the unique pre-processing step may benefit specialized recognition tasks. Conversely, Scenario C showed consistent, measurable improvement for complex and command sentences, confirming its broad effectiveness in restoring sentence structure necessary for NMT processing.

\section{Discussion}
The experimental results reveal key factors influencing the performance of cascaded S2TT for the low-resource Nepali $\rightarrow$ English translation pair. We analyze the effectiveness of component-level refinements, quantify error propagation, evaluate the Punctuation Restoration Module (PRM), and validate findings through human assessment.

\subsection{Validation of Component-Level Refinements}
\paragraph{ASR Performance:}The fine-tuned \textbf{Wav2Vec2 XLS-R} model achieved a low CER (2.72\%), representing a significant improvement over previous state-of-the-art supervised models on the OpenSLR-54 dataset. This success is primarily attributed to transfer learning. The Wav2Vec2 model, pre-trained on vast amounts of unlabelled speech data, quickly adapted its generalized features to limited Nepali data. This can also be attributed to the fact that, Nepali is very similar to Hindi, and Wav2Vec2 has been pre-trained on large amounts of unlabeled Hindi data as well, effectively mitigitaing the effects of data scarcity.

\paragraph{NMT performance:}The subsequent NMT model demonstrated high translation proficiency, outperforming previously documented models on standard benchmarks (e.g., $\sim$8--9 BLEU points on FLORES-200). This confirms the success of the data curation and MarianMT fine-tuning stages, establishing the NMT component as highly capable when provided with clean, structured input, or large-scale high-quality training data.

\subsection{Quantifying Error Propagation}
The experiment isolating punctuation loss empirically confirms the necessity of the proposed PRM intervention. By comparing NMT performance on punctuated vs. unpunctuated gold-standard text, the results showed a severe performance degradation, peaking at a \textbf{11} BLEU point loss (28.39\% relative drop) on the Tatoeba test set.

This degradation occurs because punctuation marks are essential structural tokens that define sentence bondaries and syntactic relationships, which NMT models rely on for accurate translation. Their removal forces the NMT encoder to process long, coherent streams of tokens, leading to grammatical incoherence, and a copmplete loss of contextual cues (e.g., distinguishing between a statement and a question). This finding empirically validates the core premise of this resrearch: that directly addressing the loss of structural metadata (punctuation) is the most critical intervention point for improving cascaded S2TT quality.

\subsection{Analysis of End-to-End Pipeline Performance}
The S2TT evaluation on the custom test set (with a ASR WER of 37.36\%) reveals the optimal strategy for integrating the PRM:
\paragraph{Optimal Configuration (Scenario C):}Scenario C (ASR $\rightarrow$ Punctuation Restoration (without removing spaces i.e. without addressing ASR segmentation issues) $\rightarrow$ NMT) achieved the highest S2TT scores (36.38 BLEU), providing a 4.90 BLEU point gain over the direct baseline (Scenario A). This success indicates that the ASR component's internal language model, despite its high WER, was effective at generating adequate word segmentation (i.e., correct spacing between words). The PRM in Scenario C succeeded because it made the minimal effective intervention: focusing primarily on restoring the lost punctuation and sentence boundaries without disturbing the existing, reasonably correct word spacing.

\paragraph{Sub-Optimal Configuration (Scenario B):}Scenario B (ASR $\rightarrow$ Punctuation Restoration with removing spaces from input $\rightarrow$ NMT) performed sub-optimally, achieving lower score than Scenario C. This outcome demonstrates a crucial trade-off: the deliberate pre-processinng step of removing all spaces (token segmentation) and forcing the PRM to re-segment the entire utterance introduced new, cascading segmentation errors. These self-inflicted errors proved more detrimental to the downstream NMT model than the benefit gained from the restored punctuation alone. This finding suggests that for Nepali, preserving the ASR's inherent word segmentation is critical and should not be discarded in favor of concurrent text-to-text re-segmentation unless the PRM is highly trained on a large-scale high-quality and diverse dataset for that specific task.

\subsection{Validation by Human Assessment}
The results from the human evaluation strongly corroborate the automatic metrics. \textbf{Scenario C} consistently scored the highest in both \textbf{Fluency (3.804)} and \textbf{Adequacy (3.673)}. This correlation validates that the \textbf{4.90} BLEU gain is not merely a statistical artifact but translates directly into a perceivably higher-quality, more readable, and meaning-preserving translation for the end-user. This is further supported by a high degree of consensus among evaluators, with a Krippendorff's Alpha ($\alpha$) of 0.748 for fluency and 0.785 for adequacy across the optimal configuration.

The human breakdown also highlighted specific structural benefits. While \textbf{Scenario B} showed a surprising advantage in case of \textbf{Named Entities} (suggesting value in its forced segmentation for specific tasks), Scenario C's superior performance in \textbf{Complex sentences} and \textbf{Commands} confirms its broader robustness in restoring the sentence structure necessary for NMT.

\subsection{Primary Contribution and Future Work}
\subsubsection{Primary Contribution}
The primary contribution of this work is the empirical validation and successful deployment of a Punctuation Restoration Module as a targeted intermediary step in a low-resource S2TT pipeline. We demonstrated that for Nepali, this strategy provides a robust and significant performance gain (+4.90 BLEU) by mitigating the most severe form of ASR noise -- the loss of structural cues -- thereby establishing a new, optimized baseline for Nepali-to-English S2TT.
\subsection*{Limitations and Future Work}
The main limitations stem from the cascaded nature of the system and high-quality data scarcity.
\begin{itemize}
  \item \textbf{Cascaded Error Floor:} The system's performance ceiling is ultimately limited by the \textbf{37.36\%} WER of the ASR output on the custom test set. Future work should focus on integrating the PRM and NMT into a single, end-to-end Speech-to-Text Translation model to allow for joint training and attention, potentially overcoming cascaded error propagation.
  \item \textbf{PRM Robustness:} The failure of Scenario B highlights the need for a more robust PRM capable of handling concurrent punctuaion and word segmentation. Future work should explore training the PRM on deliberately noisy, unsegmented data to improve its generalization across varying ASR output styles.
  \item \textbf{Data Generalization:} The training data for all components remains relatively narrow. Future efforts must prioritize the creation of larger, more diverse, and domain-spanning Nepali speech and parallel text corpora. 
\end{itemize}

\section{Conclusion}
This research successfully addressed the critical challenge of structural noise propagation in the Nepali-to-English cascaded S2TT pipeline, a pervasive issue for low-resource languages that lack sophisticated end-to-end models.

We first established the necessity of the intervention by empirically demonstrating that the removal of punctuation alone resulted in a substantial performance degradation, causing a $\sim$ 6 BLEU point loss on standard NMT evaluation sets.

To mitigate this core problem, we proposed and validated a Punctuation Restoration Module (PRM) as a crucial intermediate component. The optimized pipeline configuration (Scenario C), which deployed the PRM on the raw ASR output to restore structural cues, yielded a significant performance increase, achieving a \textbf{4.90} BLEU point gain over the direct ASR-to-NMT baseline. This result was further validated by human evaluation, with the optimized system scoring highest in both \textbf{Adequacy (3.673)} and \textbf{Fluency (3.804)}, reinforced by inter-rater reliability ($\alpha \geq 0.72$) for all scenarios.

The primary contribution of this work is the empirical proof that a targeted, text-to-text intervention, specifically, punctuation restoration, is an extremely effective strategy for overcoming noise inherent to low-resource ASR and significantly improving the overall quality of cascaded S2TT. This work establishes an optimized baseline for Nepali-to-English S2TT and provides a validated architectural blueprint for future research in similar low-resource language pairs.

Future work should focus on integrating the PRM and NMT into a single end-to-end Speech-to-Text Translation model to overcome the remaining limitations of cascaded error propagation and explore training a more robust PRM to concurrently handle both punctuation and word segmentation errors.

\bibliographystyle{ACM-Reference-Format}
\bibliography{citations}










\end{document}